\documentclass{ecai} 

\usepackage{latexsym}
\usepackage{amssymb}
\usepackage{amsmath}
\usepackage{amsthm}
\usepackage{booktabs}
\usepackage{enumitem}
\usepackage{graphicx}
\usepackage{color}

\newcommand{\BibTeX}{B\kern-.05em{\sc i\kern-.025em b}\kern-.08em\TeX}

\begin{document}


\begin{frontmatter}


\paperid{123} 


\title{Deep Learning based 3D Volume Correlation for Additive Manufacturing Using High-Resolution Industrial X-ray Computed Tomography}


\author[A]{\fnms{Keerthana}~\snm{Chand}\orcid{0009-0000-5748-2264}\thanks{Corresponding Author. Email: keerthana.chand@bam.de.}}
\author[A]{\fnms{Tobias}~\snm{Fritsch}\orcid{0000-0002-0472-2874}}
\author[A]{\fnms{Bardia}~\snm{Hejazi}\orcid{0009-0000-5748-2264}}
\author[A]{\fnms{Konstantin}~\snm{Poka}\orcid{0000-0001-7340-8822}}
\author[A,B]{\fnms{Giovanni}~\snm{Bruno}\orcid{0000-0001-9632-3960}} 

\address[A]{Bundesanstalt Für Materialforschung Und -Prüfung (BAM, Federal Institute for Materials Research and Testing), Berlin, Germany}
\address[B]{Institute of Physics and Astronomy, University of Potsdam, Potsdam, Germany}


\begin{abstract}

Quality control in additive manufacturing (AM) is vital for industrial applications in areas such as the automotive, medical and aerospace sectors. Geometric inaccuracies caused by shrinkage and deformations can compromise the life and performance of additively manufactured components. Such deviations can be quantified using Digital Volume Correlation (DVC), which compares the computer-aided design (CAD) model with the X-ray Computed Tomography (XCT) geometry of the components produced. However, accurate registration between the two modalities is challenging due to the absence of a ground truth or reference deformation field. In addition, the extremely large data size of high-resolution XCT volumes makes computation difficult. In this work, we present a deep learning-based approach for estimating voxel-wise deformations between CAD and XCT volumes. Our method uses a dynamic patch-based processing strategy to handle high-resolution volumes. In addition to the Dice Score, we introduce a Binary Difference Map (BDM) that quantifies voxel-wise mismatches between binarized CAD and XCT volumes to evaluate the accuracy of the registration. Our approach shows a 9.2\% improvement in the Dice Score and a 9.9\% improvement in the voxel match rate compared to classic DVC methods, while reducing the interaction time from days to minutes. This work sets the foundation for deep learning-based DVC methods to generate compensation meshes that can then be used in closed-loop correlations during the AM production process. Such a system would be of great interest to industries since the manufacturing process will become more reliable and efficient, saving time and material.

\end{abstract}

\end{frontmatter}


\section{Introduction}
Additive Manufacturing (AM) is revolutionizing the production of complex structures, which are beneficial especially in high impact sectors such as the aerospace, biomedical and automotive industries~\citep{Hamza2025,Fidan2023}. Compared to traditional manufacturing technologies, AM enables greater geometric freedom, better customization in small series and improved material efficiency~\citep{Yang2018}. However, dimension accuracy and mechanical reliability remain key challenges limiting widespread usage of AM technologies~\citep{Yadollahi2017,Islam2013AnIO}. Additively fabricated structures are often affected by geometric inaccuracies such as shrinkage and local deformations~\citep{FriedenTempleton2024,Paul2014}. These deviations are mainly caused by factors such as uneven thermal gradients and residual stress during the manufacturing process and their release after removing the structure from the build platform~\citep{Li2018,Gao2024}. Such deviations can be quantified using Digital Volume Correlation (DVC) between the actual fabricated structures and the nominal design represented by CAD~\citep{Bay1999}.

DVC is widely used in AM to evaluate displacement fields and measure strain, typically by comparing X-ray Computed Tomography (XCT) scans during in-situ testing, e.g. under different load conditions~\citep{Pan2012, Li2023}. These correlations between XCT scans of the same structure under different conditions correlations work well due to the high similarity in intensity and structure between the paired volumes. However, applying the existing methods for DVC between a nominal CAD and XCT geometry of the fabricated structure is challenging due to low intensity correlation between the two modalities. 

Chand et al. used SPAM (Software for Practical Analysis of Materials), for deformation estimation and reported the limiations of such approaches for our specific use case~\citep{Chand2025, Stamati2020}. SPAM performs DVC by iteratively optimizing local correlation windows (nodes) across 3D volumes to estimate displacement fields between two volumes. Such traditional methods struggle with limited spatial resolution because they rely on local correlation windows. These windows often fail to capture fine deformations such as warping, shrinkage, or surface roughness. Traditional DVC methods also require iterative and exhaustive search using a multiscale strategy across 3D volumes, which makes them computationally expensive~\citep{Leclerc2010}. This issue is especially pronounced for high-resolution datasets with micrometer-scale voxel sizes (e.g., 10 µm or smaller).  As a result, there exists no established method for performing voxel-level deformation estimation between CAD and XCT data at high resolutions. This remains a strong limitation for quantifying distortions and implementing closed-loop quality control in AM~\citep{Jadayel2020}. Moreover, this mismatch complicates the correlation of process-induced pore locations identified using XCT with in-situ monitoring signals, thereby limiting process understanding~\citep{Oster2024}.

To address these challenges in traditional DVC, this work proposes the first application of an unsupervised deep learning (VoxelMorph) based model for DVC on XCT data in AM quality control. Deep learning methods can estimate displacement in a single inference pass, significantly reducing computational time. To improve spatial resolution, we introduce a dynamic patch-based strategy using the HDF5 file format, enabling training at high resolution. 

We test our approach on Triply Periodic Minimal Surface (TPMS) structures with varying the level-set parameter C, creating a diverse yet representative
set of deformation in AM that can be obtained using XCT scans. For quantitative performance evaluation, we introduce the Binary Difference Map (BDM) as well as use the Dice Score to assess alignment accuracy. Based on these quality matrices, we also compared our deep learning approach to the traditional SPAM method and showed improvement in DVC performance.


\section{Method}

\begin{figure*}[h]
\centering
\includegraphics[width=15cm]{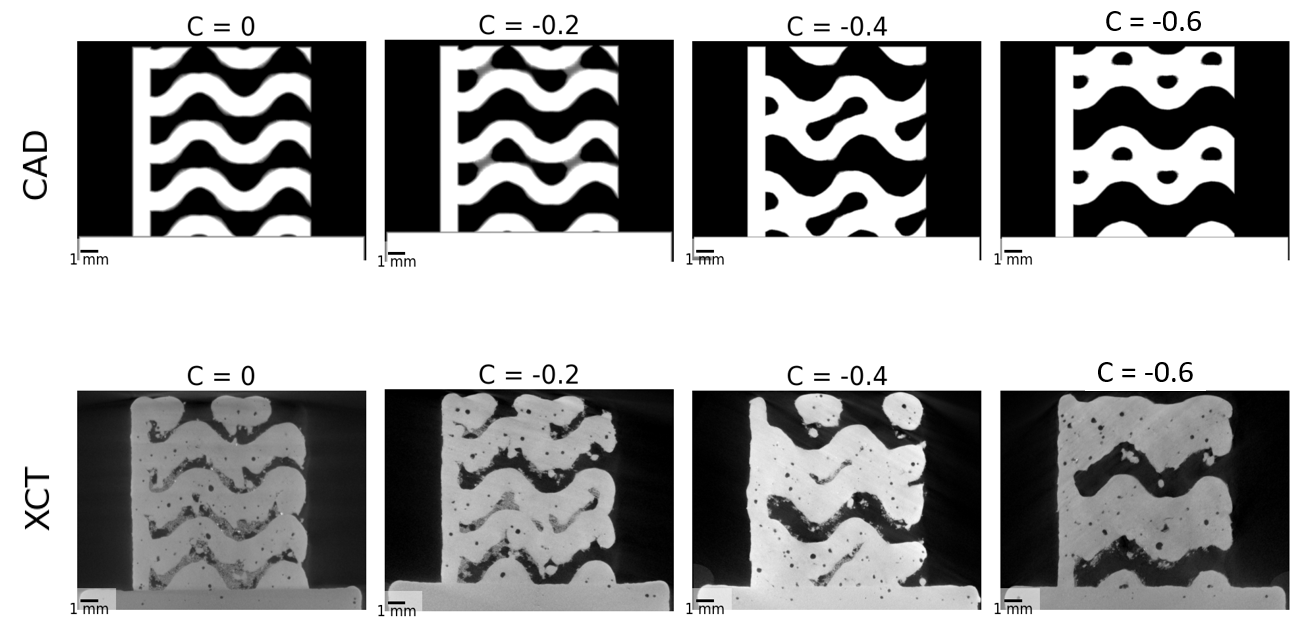}
\caption{CAD models and corresponding XCT scan cross-sections for TPMS samples with different level-set parameters C.}
\label{fig:TPMSstructure}
\end{figure*}

\begin{figure*}[h]
\centering
\includegraphics[width=16.5cm]{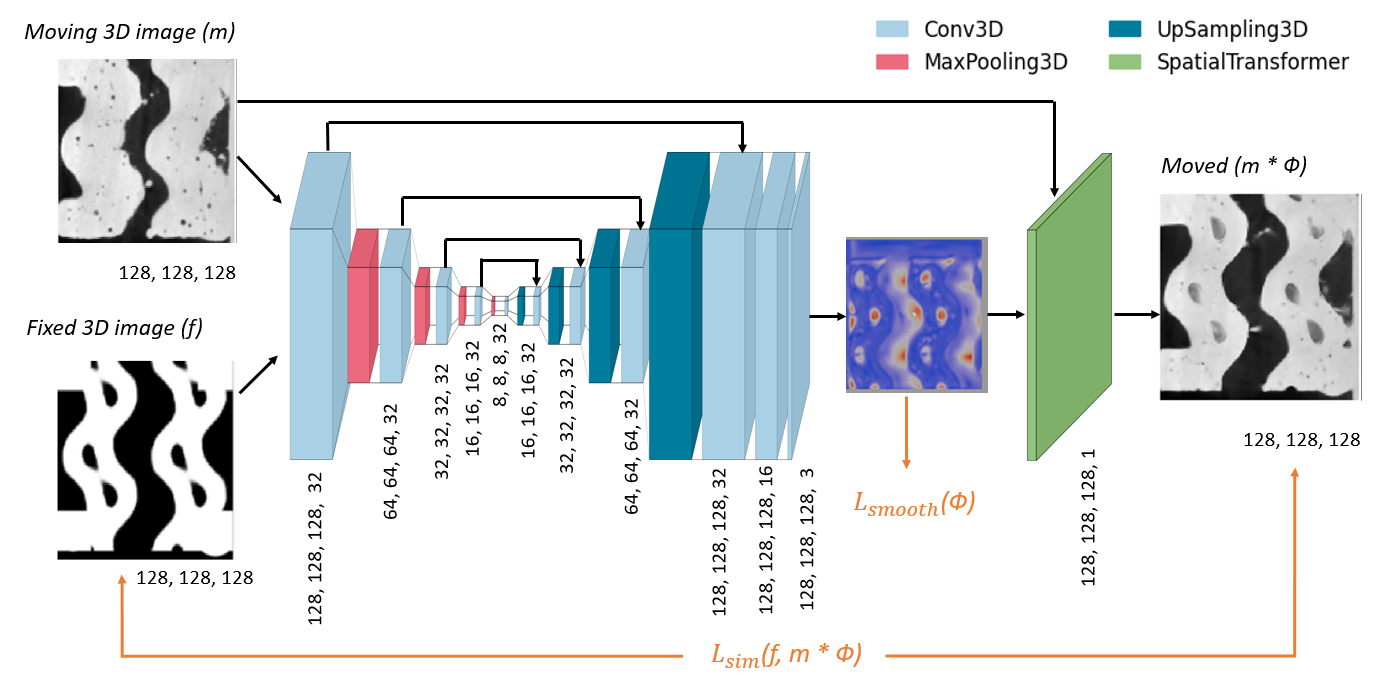}
\caption{Overview of VoxelMorph-based 3D registration framework.}
\label{fig:model}
\end{figure*}

\subsection{Data Acquisition and Preparation}

TPMS structures were selected as benchmark geometries in this study due to their complex internal structure, which resembles typical porous structures found in industrial applications~\citep{Feng2022-bk,Chouhan2023}. We hypothesize that the complex geometry of TPMS structures, combined with their thin walls (0.5 mm), leads to distortion and results in rich deformation fields, which make them well suited for training deep learning models. Moreover, by varying the level-set parameter C, which controls the iso-surface threshold in the implicit TPMS equation, the porosity of the structure and hence the deformation characteristics, can be systematically varied~\citep{Al-Ketan2018-dq,Chand2025}. This enables the generation of a large and representative training dataset essential for robust deep learning models~\citep{Yang2021}. 

Seven gyroid-based TPMS samples with varying porosity levels were designed using a level-set approximation approach. The level-set parameter C was varied from 0 to \textbf{$-$} 0.6 in steps of \textbf{$-$}0.1~\citep{Al-Ketan2018-dq,Chand2025}. Each sample had a fixed wall thickness of 0.5~mm and an overall size of $5 \times 5 \times 5$ mm³. The first row of Fig. \ref{fig:TPMSstructure} shows example CAD and XCT volume reconstruction slices of the designed TPMS structures with the level-set parameters C = 0, \textbf{$-$}0.2, \textbf{$-$}0.4 and \textbf{$-$}0.6, respectively. The images show that increasing the magnitude of C leads to a shift in the distance between the walls, resulting in a change in porosity. Spherical features of varying diameters and a 0.5~mm wall were added to the bare TPMS structures to support linear registration. Additionally, the TPMS structures were fabricated on an $8 \times 8 \times 8$ mm³ base plate, which served as the reference for linear registration~\citep{Chand2025}. The structures were fabricated using the Powder Bed Fusion with Laser Beam of Metals (PBF-LB/M) process on an EOS M 300–4 system with AlSi10Mg powder. Printing parameters included: a layer thickness of 60 $\mu$m, a laser power of 370 W for infill and 320 W for contours. The build platform was preheated to 165 °C. A bidirectional stripe scanning strategy was applied with a 47° rotation per layer.

The fabricated structures were scanned using a diondo dxMax dual-source laboratory cone beam XCT system. XCT scanning parameters were the same across all seven samples: 60 kV acceleration voltage, 167 $\mu$A target current, a 1.0 mm aluminum filter, and 1800 ms exposure time over 3900 projections. A reconstructed voxel size of 5 $\mu$m was achieved. The second row of Fig. \ref{fig:TPMSstructure} shows example slices of the reconstructed XCT volumes for different values of the level-set parameter C. 

\subsection{Data Preprocessing}

The XCT scans of all 7 samples were downsampled by a factor 2 to achieve faster preprocessing. The CAD models of the TPMS structures were voxelized and the XCT data were manually registered to the reference CAD volume using VGStudio Max software Version 2023.2.1 using the base plate as reference~\citep{Chand2025}. Powder residues and background noise were removed using morphological cleaning. This process consisting of Otsu thresholding, connected component analysis, erosion, and opening, in order to isolate the largest component, which corresponds to the fabricated structure. Subsequently, a pore-filling algorithm was applied to compensate for keyhole porosities which are caused by the AM process. Then, the corresponding region of interest (ROI) was extracted from both the CAD and cleaned XCT volumes. A common target shape of 816 × 592 × 816 voxels was used to ensure compatibility with the neural network input. Min–max normalization was then applied to both volumes to facilitate effective learning~\citep{CabelloSolorzano2023}. Among the seven TPMS samples, four (with level-set parameters C = 0, -0.2, -0.3, and -0.5) were allocated for training, two (C = -0.1 and -0.4) for validation, and one (C = -0.6) for testing. This corresponds to a 60 / 20 / 20 split ratio. The preprocessed volumes were then organized into training (24.4 GB), validation (11.7 GB), and test (5.8 GB) splits. The volumes are stored in HDF5 file format, which contains corresponding pairs of preprocessed XCT and CAD volumes, along with sample-wise metadata for efficient access during model training~\citep{Folk2011}. HDF5 enables dynamic data loading and patch extraction during the model's workflow, avoiding the need to pre-store large numbers of subvolumes and simplifying the handling of large 3D datasets. All preprocessing and data handling steps were implemented in Python version 3.9. Libraries used included scikit-image and SciPy for image processing and morphological operations, and h5py for HDF5 storage.

\subsection{VoxelMorph-Based Unsupervised Model for 3D Image Registration}

The unsupervised deep learning model for 3D image registration is based on the VoxelMorph architecture. It was implemented in TensorFlow using the Voxelmorph and Neurite libraries~\citep{Balakrishnan2019,dalca2018anatomical}. VoxelMorph was selected for its unsupervised learning framework and demonstrated effectiveness in estimating non-linear deformation fields between volumetric datasets in medical imaging~\citep{Balakrishnan2019}. Its ability to learn smooth and dense displacement fields without requiring ground-truth annotations makes it a suitable choice for industrial XCT applications, where such labels are typically unavailable. 

The model takes as input a pair of volumes: a moving XCT volume and a fixed CAD volume. It outputs both the warped (registered) XCT volume and the corresponding displacement field. The output displacement field has the same spatial dimensions as the input volume, with three dimensions representing voxel-wise displacements along the x, y, and z axes.

The deep learning model has a 3D encoder–decoder architecture with skip connections, as illustrated in Fig.~\ref{fig:model}. The input volumes are concatenated into a 2-channel 3D image. The encoder is comprised of four levels of 3D convolutional layers with feature dimensions [32, 32, 32, 32]. Each convolutional layer is followed by a LeakyReLU activation and a 3D max pooling operation to progressively reduce the spatial resolution. The encoder captures hierarchical features between the XCT and CAD volumes and maps them into a lower-dimensional latent space. 

The decoder reconstructs the deformation field through six convolutional blocks with output channels [32, 32, 32, 32, 32, 16]. At each of the first four levels, UpSampling3D is used to progressively increase spatial resolution until it matches that of the input volume. Skip connections from corresponding encoder levels are concatenated with decoder layers to directly transfer learned features and support the decoder in generating the deformation field. Each of these blocks is followed by  a 3D convolution and LeakyReLU activation. Once the full input resolution (128 × 128 × 128) is reached, two additional convolutional layers are applied. These layers use only a convolution and LeakyReLU activation, without any further upsampling. This refined feature map is then passed to a final convolutional layer to predict the 3D displacement field.

A final 3D convolutional layer produces a three-channel displacement field representing voxel-wise shifts in x, y, and z directions. This displacement field is applied to the moving image using a Spatial Transformer layer to obtain the registered XCT volume. 

During training, random patches were dynamically extracted from different samples and spatial locations within the preprocessed volumes using a Python generator. This approach reduces memory usage and improves the model’s generalization capacity. Patch-wise training also allows the model to focus on fine local deformations rather than overfitting to global geometry of the structure~\citep{Lotz2016}. The model was optimized using Normalized Cross-Correlation (NCC) loss for image similarity and L2 gradient loss for regularizing the smoothness of the displacement field. A weighting factor of $\lambda$ = 0.05 was used~\citep{Balakrishnan2019}.

The model was trained using the Adam optimizer with a learning rate of 0.001 for 200 epochs, with 5 steps per epoch and a batch size of 8~\citep{adam}. Validation was performed using a batch size of 4, and the learning rate was adaptively reduced using ReduceLROnPlateau based on validation loss. The model was trained on a high-performance computing (HPC) system using a single NVIDIA A100 GPU. The training took approximately 8 hours. The model training showed stable convergence without signs of overfitting. 

For inference on the test data, the trained VoxelMorph model employs a dynamic patch-wise strategy with Gaussian-weighted blending. The weight mask assigns higher importance to the central voxels of each patch, ensuring smooth transitions at the patch borders~\citep{Christ2016}. The Python generator extracts patches of size 128 × 128 × 128, with a stride of 64 × 64 × 64, and feeds them into the model sequentially. Each predicted moved patch and displacement field is reconstructed into a global volume using Gaussian-weighted blending.  

\subsection{Performance Evaluation}

To evaluate the accuracy of the VoxelMorph model, both quantitative metrics and qualitative visualizations were used. We also report the inference time taken for model deployment on all datasets used. Two primary metrics were used to quantify the results: the Dice Score~\citep{Dice1945-ko} and the BDM. Dice Score was used to measure the spatial overlap between the volumes before and after non-linear registration following prior works in image registration~\citep{Balakrishnan2019,Fehr2016}. The Dice Score was computed after binarizing the volumes using global Otsu thresholding. We computed BDM, inspired by binary error maps used in segmentation tasks but adapted to our use case~\citep{Ronneberger2015}.

Inorder to calculate BDM, both XCT and CAD volumes were binarized using global Otsu thresholding. We then compute a voxel-wise difference restricted to the union of the foreground regions of the binarized XCT and CAD volume:

\begin{equation}
\text{BDM} = \text{XCT}_{\text{bin}} - \text{CAD}_{\text{bin}}
\end{equation}

This results in three possible values per voxel:
\begin{itemize}
    \item 0 : Voxel is present in both XCT and CAD
    \item +1 : Voxel is present in XCT but not in CAD 
    \item –1 : Voxel is present in CAD but not in XCT 
\end{itemize}

Based on the BDM map, we compute three key metrics, considering only voxels within the union of the binarized XCT and CAD volumes:

\begin{equation}
\text{BDM}_{0} = \frac{\text{Number of voxels with 0} }{\text{XCT}_{\text{bin}} \cup \text{CAD}_{\text{bin}}} \times 100\%
\end{equation}

\begin{equation}
\text{BDM}_{+1} = \frac{\text{Number of voxels with } +1}{\text{XCT}_{\text{bin}} \cup \text{CAD}_{\text{bin}}} \times 100\%
\end{equation}

\begin{equation}
\text{BDM}_{-1} = \frac{\text{Number of voxels with } -1}{\text{XCT}_{\text{bin}} \cup \text{CAD}_{\text{bin}}} \times 100\%
\end{equation}

These metrics help to characterize the types of misalignment. In detail, $\text{BDM}_{0}$ quantifies the percentage of voxels where the XCT and CAD volumes agree, representing regions where the fabricated structure matches the CAD design. $\text{BDM}_{+1}$ measures voxels that are present in the XCT volume but not in the CAD model. This suggests excess material due to factors such as deformation, feature loss due to fabrication limitations, partially sintered powder or surface roughness. $\text{BDM}_{-1}$ corresponds to voxels that are present in the CAD model but missing in the XCT volume. This indicates material loss due to factors such as structural breakage, shrinkage or surface roughness. 

The performance of the VoxelMorph model was also compared against SPAM. For both methods, SPAM and VoxelMorph, the Dice Score and BDM were computed on the same test sample to ensure a fair comparison. 

The CAD, XCT, and registered XCT volumes were visualized using orthogonal slice views to qualitatively evaluate the alignment performance. Additionally, the magnitude and direction of the displacement field were plotted to better understand the deformation patterns. Matplotlib and PyVista were mainly used for the visualizations.


\section{Results and Discussion}
We present results in terms of registration accuracy, deformation field visualization, and comparison with the baseline method, SPAM. We also discuss the strengths and limitations of the model in this section.

\subsection{Evaluation Metrics and displacement field analysis}

\begin{table*}[h]
\caption{Dice Scores before and after registration and runtime for all TPMS structures.}
\centering
\begin{tabular}{lcccc} 
\toprule
\textbf{Dataset} & \textbf{Split} & \textbf{Dice Before (\%)} & \textbf{Dice After (\%)} & \textbf{Time (min)} \\
\midrule
c = 0  & Train      & 82.63 & 95.92 & 6.25 \\
c = -0.2  & Train      & 83.01 & 96.86 & 6.07 \\
c = -0.3  & Train      & 84.41 & 97.66 & 6.15 \\
c = -0.5  & Train      & 83.06 & 97.09 & 6.12 \\
c = -0.1  & Validation & 81.30 & 96.09 & 5.95 \\
c = -0.4  & Validation & 83.71 & 97.50 & 5.81 \\
c = -0.6  & Test       & 82 & 94.7 & 5.81 \\
\bottomrule
\end{tabular}
\label{tab:vertical_dice_results}
\end{table*}

Table \ref{tab:vertical_dice_results} shows the Dice Score before and after non-linear registration. It also reports the run time for all TPMS structures used for the study. Results are reported for training, validation, and test datasets.

The model consistently improved the alignment between the CAD and XCT volumes across all cases, as indicated by the increased Dice Scores after registration. On the test data, the Dice Score increased from 82\% to 94.7\%. On average, the Dice Score improved from 83\% before registration to 96.6\% after registration, resulting in an average improvement of 13.6\% across all datasets. The validation and test sets also show similar trends to the training sets, confirming the model’s ability to generalize to unseen data. The average runtime for each registration was approximately 6 minutes, which included both data loading and inference time on GPU.

\begin{figure*}[h]
\centering
\includegraphics[width=13cm]{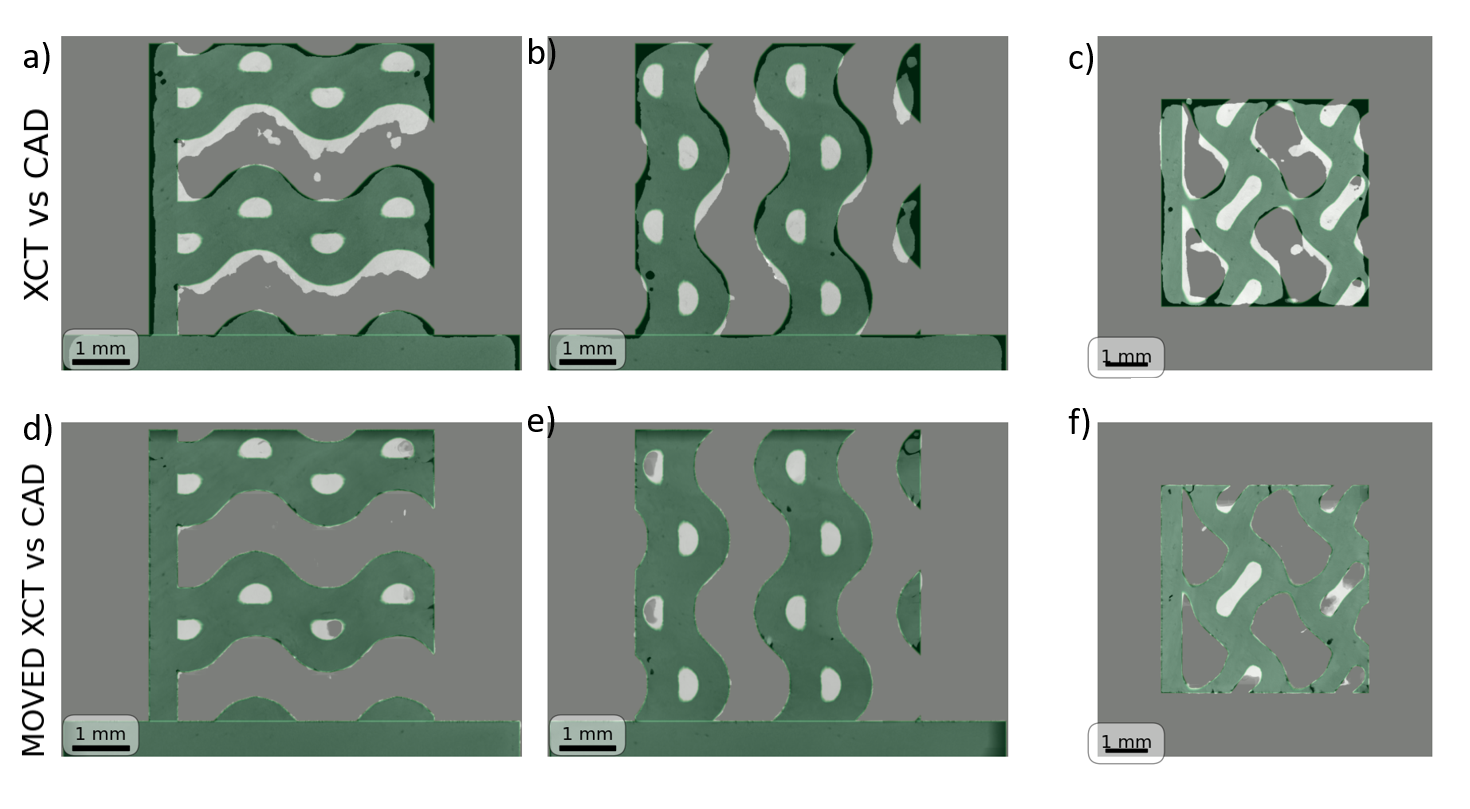}
\caption{Overlay of XCT (grey) and CAD (green) of the central slices for the sample with C = -0.6, shown before (a, b, c) and after (d, e, f) non-linear registration. Views include: (a, d) XZ plane, (b, e) YZ plane and (c, f) XY plane.}
\label{fig:ctandmovedct}
\end{figure*}

Fig.~\ref{fig:ctandmovedct} shows the overlay of the XCT (gray) and CAD (green) images before and after applying the non-linear registration on the test sample. Fig.~\ref{fig:ctandmovedct} (a, b, c) displays the central slices in the XZ, YZ, and XY planes, respectively, before the registration. Various deviations can be observed in these images. Along the structure boundaries, deviations primarily arise from material shrinkage and surface roughness. Internal structures also show deviations: larger TPMS intrinsic pores (intended as the hollow parts of the structure) appear smoother and less deformed, while smaller TPMS intrinsic pores often have significant deformation and are mostly entirely closed. This effect might be due to the resolution limitations of the PBF-LB/M machine. Notably, the down-skin surfaces in the XZ and YZ views show greater deformation than the up-skin surfaces, as the absence of supporting material during fabrication leads to increased structural instability and heat accumulation in bridge-like structures. 

Fig.~\ref{fig:ctandmovedct} (d, e, f) shows the same slices after non-linear registration. An improvement in alignment is achieved especially at the edges, downskin and larger internal structures. This result shows the model’s efficiency in compensating for shrinkage and local deformations. However, deviations remains unchanged in regions with closed TPMS intrinsic pores.

\begin{figure*}[h]
\centering
\includegraphics[width=17cm]{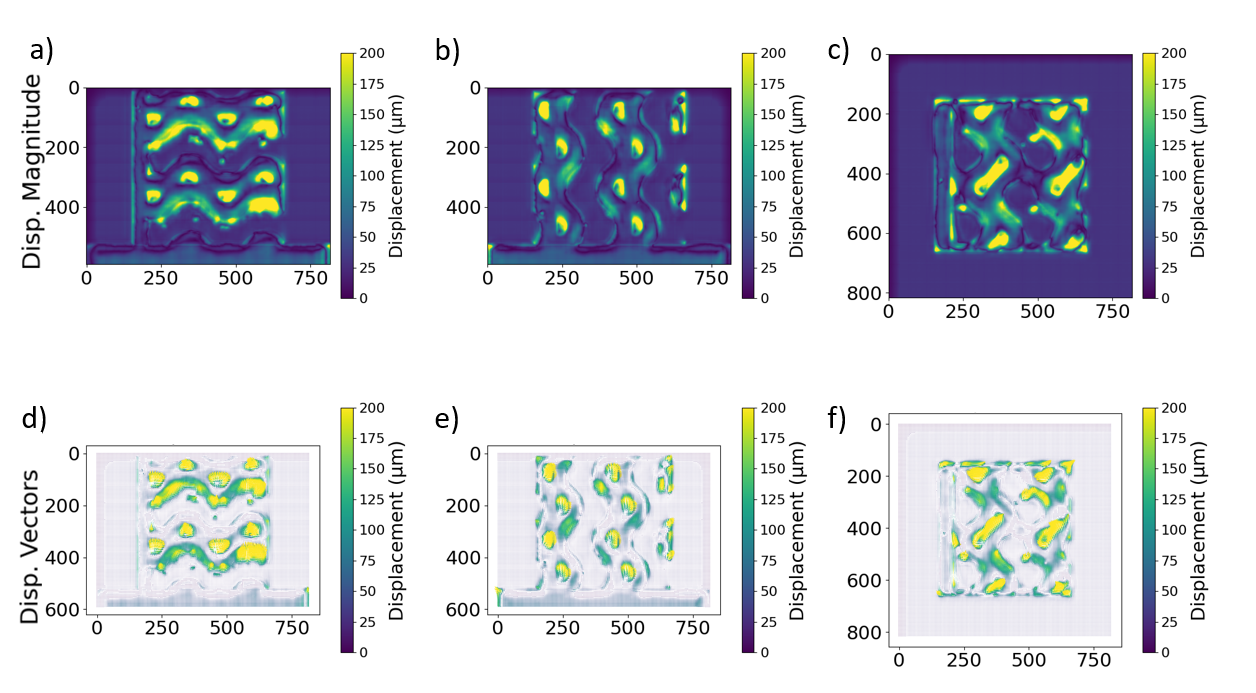}
\caption{Visualization of the deformation field estimated by the VoxelMorph model on the same slice shown in Fig.~\ref{fig:ctandmovedct}. (a, b, c) shows the magnitude of the displacement field in the central slices along the XZ, YZ, and XY planes, respectively. The bottom row (d, e, f) presents the corresponding displacement field of the same slices.}
\label{fig:disp_field}
\end{figure*}

Fig.~\ref{fig:disp_field} shows the magnitude and direction of the displacement field directed from the XCT to the CAD volumes of the same slice as Fig.~\ref{fig:ctandmovedct}. Fig.~\ref{fig:disp_field} provides insight into the nature of geometric distortions across the structure. High-magnitude displacements (yellow regions) are visible especially at edges, downskin and pore locations. However, at the pore locations, the spatial transformation fails to “open” the TPMS intrinsic pores because the XCT volume lacks structural boundaries or sufficient contrast in these regions. This is likely due to partially sintered powder or residual powder remaining after de-powdering, which the model is unable to fully compensate for. Instead, the model warps solid intensity values, resulting in a solid region despite the high deformation field. This highlights a limitation of the VoxelMorph registration approach in AM applications where features are missing in the moving image. 

\begin{figure*}[h] 
\centering
\includegraphics[width=13cm]{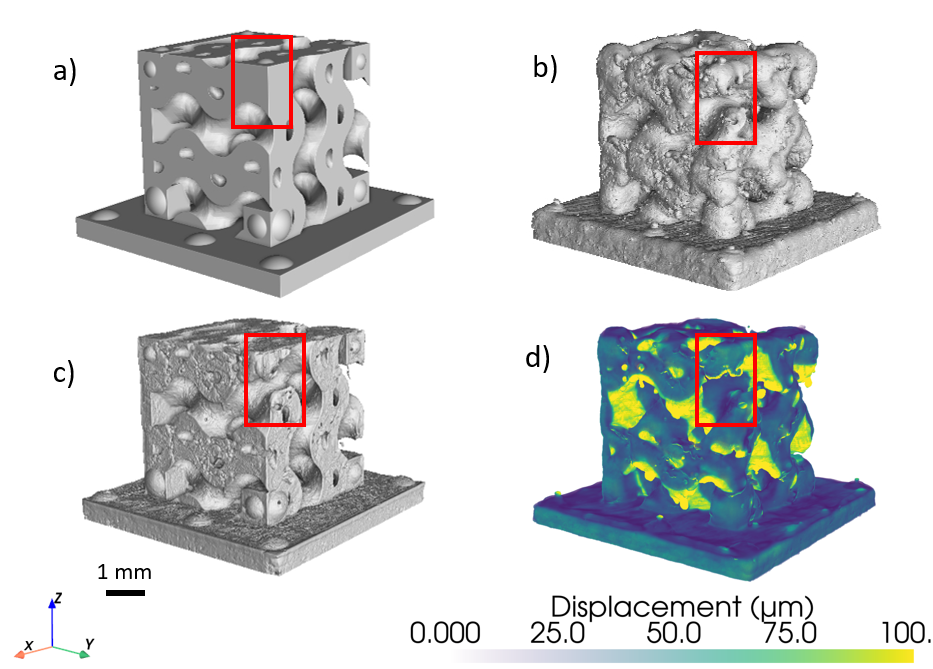}
\caption{3D visualization of registration results. a) Reference CAD model b) Original XCT volume c) Registered XCT volume aligned to the CAD geometry using the VoxelMorph model d) Masked displacement magnitude map using registered XCT volume. The red boxes highlight a region with severe structure breakage.}
\label{fig:3dvis}
\end{figure*}

Fig.~\ref{fig:3dvis} shows a 3D visualization of the registration results of the test sample, showing the CAD (Fig.~\ref{fig:3dvis}. a) and XCT (Fig.~\ref{fig:3dvis}. b) volumes, as well as the output registered XCT (Fig.~\ref{fig:3dvis}. c) and displacement magnitude (Fig.~\ref{fig:3dvis}. d) of the model. An improvement in alignment is visible in the registered XCT volume when compared to the original XCT volume. Fig.~\ref{fig:3dvis}.d maps the magnitude of displacement applied during registration, highlighting areas with high deformation (yellow regions). The red boxes indicate a region that was broken off during the fabricating process. This resulted in a physical absence of material in the XCT volume that is present in the CAD volume. The damage likely resulted from warping during fabrication, followed by a recoater collision that physically detached the region from the part. This leads to structural mismatch, which poses challenges for the registration algorithm, since the registration model attempts to align corresponding voxels (structures).

\begin{figure*}[h]
\centering
\includegraphics[width=15cm]{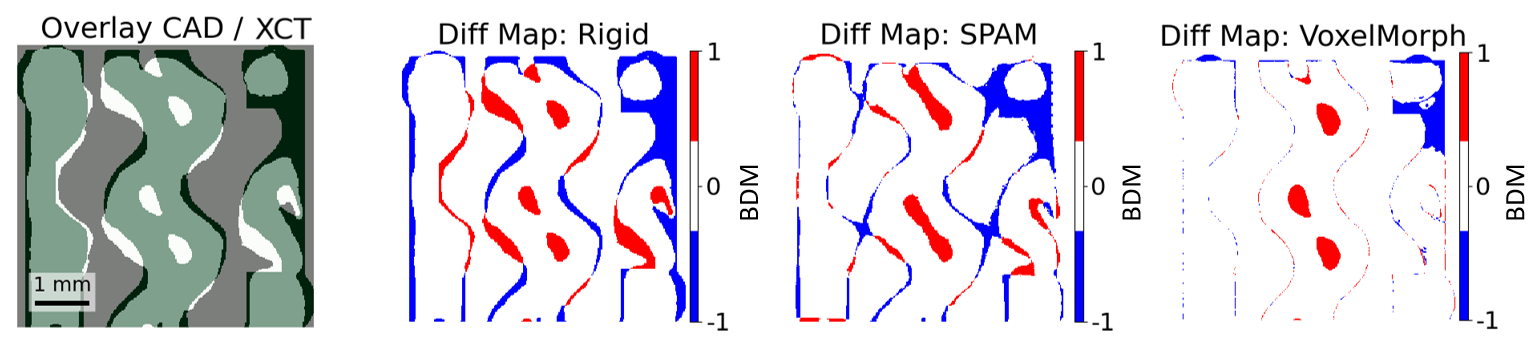}
\caption{Comparison of binary difference map (BDM) before non-linear registration, after SPAM based registration, and after VoxelMorph based registration.}
\label{fig:diff_map}
\end{figure*}

\begin{table*}[h!]
\centering
\caption{Comparison of SPAM and VoxelMorph using Dice Scores and binary difference map (BDM) before and after non-linear registration. Initial values after linear registration are common for both methods.}
\label{tab:registration_comparison}
\resizebox{\textwidth}{!}{%
\begin{tabular}{llcccc}
\toprule
\textbf{Stage} & \textbf{Method} & \textbf{$\text{BDM}_{-1}$ (CAD=1)} & \textbf{$\text{BDM}_{0}$ (Match)} & \textbf{$\text{BDM}_{+1}$ (XCT=1)} & \textbf{Dice Score} \\
\midrule
Initial (Linear Reg) & Both       & 20.50\% & 68.27\% & 11.23\% & 82\% \\
After Non-Linear Reg & SPAM       & 11.82\% & 80.01\% & 8.17\% & 85.5\%\\
                     & VoxelMorph & 2.57\% & 89.93\% & 7.5\% & 94.7\% \\
\bottomrule
\end{tabular}%
}
\end{table*}

\subsection{Comparison with SPAM} 
  
To compare the VoxelMorph model with DVC using SPAM~\citep{Chand2025}, we focus on the comparison of the BDM and the Dice Score. These metrics provide insight into how both methods capture effects such as shrinkage and local deformations. 

Table~\ref{tab:registration_comparison} shows the metrics for quantitative comparison of the registration methods. Fig.~\ref{fig:diff_map} shows the comparison of BDM before registration, after registration using SPAM, and after registration using VoxelMorph. In the difference maps, blue indicates $\text{BDM}_{-1}$, where CAD has material but XCT does not. White indicates $\text{BDM}_{0}$, i.e., matching voxels, and red indicates $\text{BDM}_{+1}$, where XCT has material but CAD does not. Initially, both SPAM and VoxelMorph start from the same rigidly registered volumes with a Dice Score of 82\% and $\text{BDM}_{0}$ of 68.27\% . After non-linear registration, VoxelMorph performs better than SPAM in terms of Dice Score (94.7\% compared to 85.5\% for SPAM) and $\text{BDM}_{0}$ (89.93\% compared to 80.01\% for SPAM). Furthermore, $\text{BDM}_{-1}$ drops significantly to 2.57\% with VoxelMorph, compared to 11.82\% in SPAM. This demonstrates the enhanced capability of VoxelMorph in compensating for local deformation present at the boundaries of the structure. SPAM fails to capture such local information due to its limited node density. Fig.~\ref{fig:diff_map} further visualizes the difference in performance between SPAM and VoxelMorph and demonstrates the advantages of the deep-learning based VoxelMorph approach over the traditional node-based SPAM framework.

Additionally, SPAM iteratively solves for the displacement field estimation problem. Such problem can take hours or even days, especially when a multiscale strategy is used on high-resolution XCT~\citep{Chand2025}. Compared to SPAM, VoxelMorph reduces the computation time from days to minutes once trained. VoxelMorph can also be scaled to higher-resolution XCT scans using the patch-based approach, whereas SPAM is limited by memory and computational demands. One major disadvantage of VoxelMorph is the need for representative training data, whereas SPAM can work on unseen data without prior training.

\subsection{Strength and limitation} 

The main strength of the VoxelMorph based approach is its ability in learning smooth and voxel-level deformation fields without being limited by node spacing. Additionally, the unsupervised learning approach of VoxelMorph is beneficial for AM, where estimating ground truth deformations is nearly impossible. Voxelmorph's patch-based implementation allows the model to scale to high resolution volumes at a lower run time when compared to traditional methods. By warping the entire moving image toward the fixed image, VoxelMorph effectively captures effects such as shrinkage and local deformations. 

A primary challenge in using the VoxelMorph method is compensating for the surface roughness as seen in Fig.~\ref{fig:diff_map}, which is highly localized and stochastic in nature, making it difficult for the model to learn a consistent pattern. Additionally, broken edges (marked with red in Fig.~\ref{fig:3dvis}) in the XCT data are a main contributor to the remaining $\text{BDM}_{-1}$ voxels in the difference map, as seen in Fig. ~\ref{fig:diff_map}, as the model cannot align missing structures in the XCT volume. Furthermore, the residual $\text{BDM}_{+1}$, where XCT shows material but CAD does not, are often caused by closed TPMS intrinsic pores in XCT due to fabricating limitations. Since there is no structural boundary in XCT to “pull apart,” even a high predicted displacement cannot recover such features. The challenges mentioned are mainly due to the physical limitations in the manufacturing process and post-processing steps such as depowdering, rather than the shortcomings of the model itself. 


\section{Conclusions and Future Work}

In this work, we presented the first deep learning-based approach for performing voxel-level 3D Digital Volume Correlation (DVC) between nominal CAD models and high-resolution X-ray Computed Tomography (XCT) volumes in Additive Manufacturing (AM) quality control. Our approach uses the unsupervised VoxelMorph model with a dynamic patch-based training strategy and HDF5-based data handling. This appraoch addresses key challenges such as modality mismatch, absence of ground truth, and handling high-resolution data in AM. Additionally, we introduced the Binary Difference Map (BDM) to provide voxel-wise match and mismatch information. The BDM complements the Dice Score in quantifying deformation.

Our main findings are:
\begin{itemize}
    \item Deep learning models can effectively learn deformation patterns such as shrinkage, local distortions, and misalignments in AM.
    \item Patch-based training using HDF5 enables memory-efficient data handling of high-resolution XCT volumes.
    \item Training on patches also helps the model focus on local information instead of the geometry of the entire volumes. This strategy enhances the model's generalization capacity.
    \item After applying our registration method on the test data, the Dice Score increased from 82\% to 94.7\%. $\text{BDM}_{0}$ (the voxel match rate) increased from 68.27\% to 89.93\%, while $\text{BDM}_{-1}$ and $\text{BDM}_{+1}$ (which quantify the voxel misalignment between CAD and XCT) were reduced. 
    \item When compared to SPAM (traditional DVC approach), the proposed method improves Dice Score by 9.2\% and significantly reduces computational time from days to minutes on the test data.
\end{itemize}

However, the model is fails in regions with broken edges, where corresponding structures are completely absent in the XCT volume. Additionally, the model cannot recover closed TPMS intrinsic pores in XCT volumes that appear as solid due to fabrication artifacts. To address these challenges, future work should incorporate uncertainty estimation methods, such as generating confidence scores for the predicted deformation fields. This is essential for improving the trustworthiness of the approach.

Additionally, we aim to explore the use of estimated deformation fields in generating compensation meshes. These meshes can be used to modify the original CAD design, supporting closed-loop corrections in the AM workflow~\citep{Jadayel2020}. This would allow improved accuracy, speed, as well as reduced material waste in AM. 

The deep learning approach can also be extended to other DVC applications involving high resolution XCT volumes, such as analysis of in-situ experiments or crack identification.

\section*{Data availability}
The data and code used in this work are available upon request.




\bibliography{mybibfile}

\end{document}